\documentclass[10pt,twocolumn,letterpaper]{article}

\usepackage{iccv}              
\usepackage{amsmath,amsfonts,bm}

\def\eqref#1{equation~\ref{#1}}

\def\1{\bm{1}}

\DeclareMathAlphabet{\mathsfit}{\encodingdefault}{\sfdefault}{m}{sl}
\SetMathAlphabet{\mathsfit}{bold}{\encodingdefault}{\sfdefault}{bx}{n}

\def\gL{{\mathcal{L}}}

\usepackage{colortbl}
\definecolor{Highlight}{HTML}{E6F1F9}
\newcommand{\chl}{\cellcolor{Highlight}}
\definecolor{RelativeGain}{HTML}{E0F1F7}

\newcommand{\beq}{\begin{equation}}
\newcommand{\eeq}{\end{equation}}
\newcommand{\beqa}{\begin{eqnarray}}
\newcommand{\eeqa}{\end{eqnarray}}

\definecolor{iccvblue}{rgb}{0.21,0.49,0.74}
\usepackage[pagebackref,breaklinks,colorlinks,allcolors=iccvblue]{hyperref}

\def\paperID{3175} 
\def\confName{ICCV}
\def\confYear{2025}

\usepackage{indentfirst}
\usepackage{float}
\usepackage{multicol}
\usepackage{multirow}
\usepackage{graphicx}
\usepackage{adjustbox}  
\usepackage{algorithm}
\usepackage{algorithmic}
\usepackage{amsmath}
\usepackage{microtype}

\begin{document}

\title{G2PDiffusion: Cross-Species Genotype-to-Phenotype Prediction via \\Evolutionary Diffusion}


\author{
    Mengdi Liu$^{1,2}$, Zhangyang Gao$^{3}$,
    Hong Chang\thanks{Corresponding author}~~$^{1,2}$,
    Stan Z. Li$^{*3}$, Shiguang Shan$^{1,2}$, Xilin Chen$^{1,2}$ \\ 
    $^1$Key Laboratory of Intelligent Information Processing of Chinese Academy of Sciences (CAS), \\ Institute of Computing Technology, CAS, China \\
    $^2$University of Chinese Academy of Sciences, China \\
    $^3$AI Lab, Research Center for Industries of the Future, Westlake University \\
}


\twocolumn[{
\renewcommand\twocolumn[1][]{#1}%
\maketitle
\begin{center}
    \centering
    \captionsetup{type=figure}
    \vspace{-2em}
    \includegraphics[width=0.9\textwidth]{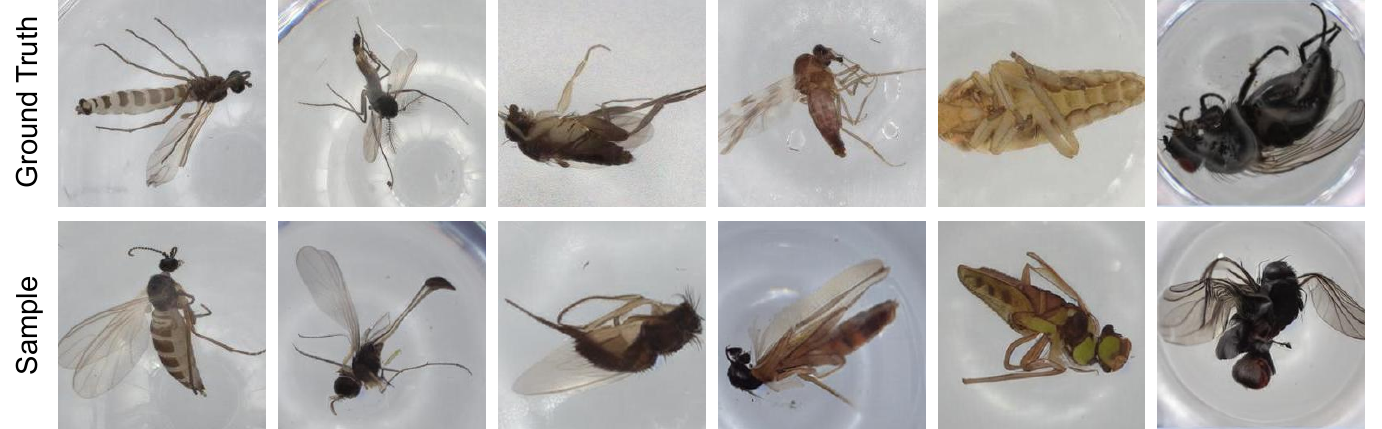}
    \vspace{-0.5em}
    \captionof{figure}{Ground truth images (top row) and generated images conditioning on DNA (bottom row).}
    \label{fig:1_overview_result}
\end{center}
}]

\begin{abstract}

\vspace{-2mm}

Understanding how genes influence phenotype across species is a fundamental challenge in genetic engineering, which will facilitate advances in various fields such as crop breeding, conservation biology, and personalized medicine. 
However, current phenotype prediction models are limited to individual species and expensive phenotype labeling process, making the genotype-to-phenotype prediction a highly domain-dependent and data-scarce problem. To this end, we suggest taking images as morphological proxies, facilitating cross-species generalization through large-scale multimodal pretraining. We propose the first genotype-to-phenotype diffusion model (\textbf{G2PDiffusion}) that generates morphological images from DNA considering two critical evolutionary signals, i.e., multiple sequence alignments (MSA) and environmental contexts. The model contains three novel components: 1) a MSA retrieval engine that identifies conserved and co-evolutionary patterns; 2) an environment-aware MSA conditional encoder that effectively models complex genotype-environment interactions; and 3) an adaptive phenomic alignment module to improve genotype-phenotype consistency. Extensive experiments show that integrating evolutionary signals with environmental context enriches the model's understanding of phenotype variability across species, thereby offering a valuable and promising exploration into advanced AI-assisted genomic analysis.

\end{abstract}
\section{Introduction}
\label{sec:intro}

One of the fundamental biology challenges is understanding how genes interact with environmental factors to determine phenotype \cite{lehner2013genotype}, which has profound implications for crop breeding \cite{araus2014field,danilevicz2022plant}, disease resistance \cite{weatherall2001phenotype}, and personalized therapeutics \cite{nebert2012personalized}. Phenotypes can be physiological, morphological, and behavioral, such as the resistance to toxins, wing shape, and foraging behavior. This paper focuses on morphological phenotypes, aiming to understand how genes influence phenotypes, how species evolve under natural selection, and how phenotypic diversity is formed.

\begin{figure}[h]
    \centering
    \includegraphics[width=1.0\linewidth]{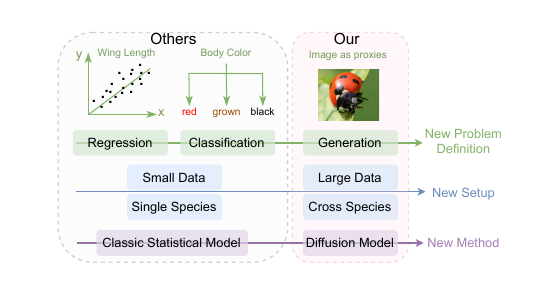}
    \vspace{-6mm}
    \caption{G2PDiffusion generates morphological images using advanced diffusion model and cross-species large data.}
    \label{fig:cover_idea}
\end{figure}

Conventional genotype-to-phenotype prediction usually relies on statistical methods such as genome-wide association studies (GWAS) \cite{uffelmann2021genome, tam2019benefits, visscher2012five,gallagher2018post,harris2024genome} and quantitative trait locus (QTL) mapping \cite{kearsey1998qtl,mccough1995qtl,korstanje2002qtl}. Recent works \cite{wang2024lstm, andreu2024phenolinker, yelmen2024interpreting,dingemans2023phenoscore} apply deep learning models to decode the intricate genotype-phenotype interactions. 
However, the existing approaches are limited to individual species due to the expensive phenotype labeling process.  As the phenotypic features are located in high-dimensional space and measured using specified equipment, labeling large populations of individuals requires intensive effort. The labeling cost is even more enormous when studying complex genotype-phenotype relationships across different species.
To break through the limitation induced by high-dimensional phenotype space, we propose to solve the problem from a novel perspective. As shown in Fig.~\ref{fig:cover_idea}, we suggest taking images as phenotypic proxies and formulating the genotype-to-phenotype prediction problem as conditional image generation. By learning from millions of DNA-image pairs across diverse taxa, our framework facilitates efficient and scalable cross-species genotype-to-phenotype prediction.

\vspace{-3mm}
\vspace{-3mm}

We propose the first genotype-to-phenotype diffusion model (\textbf{G2PDiffusion}) that generates morphological images from DNA considering two evolutionary signals, i.e., multiple sequence alignments (MSA) and environmental contexts. MSA identify evolutionary conserved and variable regions in DNA sequences, revealing genetic variations across individuals or species that contribute to morphological diversity. However, phenotypic traits are not solely determined by genotype; they are also influenced by external factors such as climate, food sources, and social interactions. Considering these influences, we take latitude and longitude as environmental factors. Both MSA and environmental contexts are regarded as evolutionary signals, enhancing the accuracy and realism of phenotype prediction.

G2PDiffusion contains three novel components: a MSA retrieval engine, an environment-aware MSA conditioner, and a dynamic genotype-phenotype aligner. Firstly, the MSA engine retrieves DNA alignments from an external database to identify evolutionarily conserved and variable sequence regions. Secondly, the retrieved MSA and environmental contexts are fed into a conditional encoder, which leverages novel MSA attention modules to capture genotype-environment (GxE) interactions. Then, we build a diffusion model conditioned on the GxE representation to generate images capturing morphological features. During each denoising step, a dynamic phenomic alignment module is employed to refine phenotypic representations.

We rigorously assess the performance of our proposed approach by comparing it with competitive baselines across diverse species under both seen and unseen conditions. We employ a range of quantitative metrics—including alignment scores, success rates, and phenotype embedding similarities—to evaluate the accuracy, biological relevance, and consistency of the generated images with the underlying genotype information. Extensive experiments demonstrate that our method not only significantly outperforms traditional models but also effectively captures the intricate genotype-environment interactions, thereby establishing its robustness and generalizability for cross-species phenotype prediction.

In summary, our \textbf{contributions} are as follows: 
\begin{itemize}
    \item We redefine the genotype-to-phenotype prediction problem as a conditional image generation, offering a novel solution to address the challenges of modeling complex environment-genotype-phenotype interactions.
    \item We propose G2PDiffusion, a first-of-its-kind diffusion model for genotype-to-phenotype prediction, where a novel evolution-aware conditional mechanism and a dynamic alignment module are proposed.
    \item G2PDiffusion can predict phenotype from genotype with high accuracy and consistency (Figure \ref{fig:1_overview_result}), offering a valuable exploration into AI-assisted genomic analysis.
    
\end{itemize}

\section{Related Works}
\label{sec:related_works}

\begin{figure*}[t]
    \centering
    \includegraphics[width=1\linewidth]{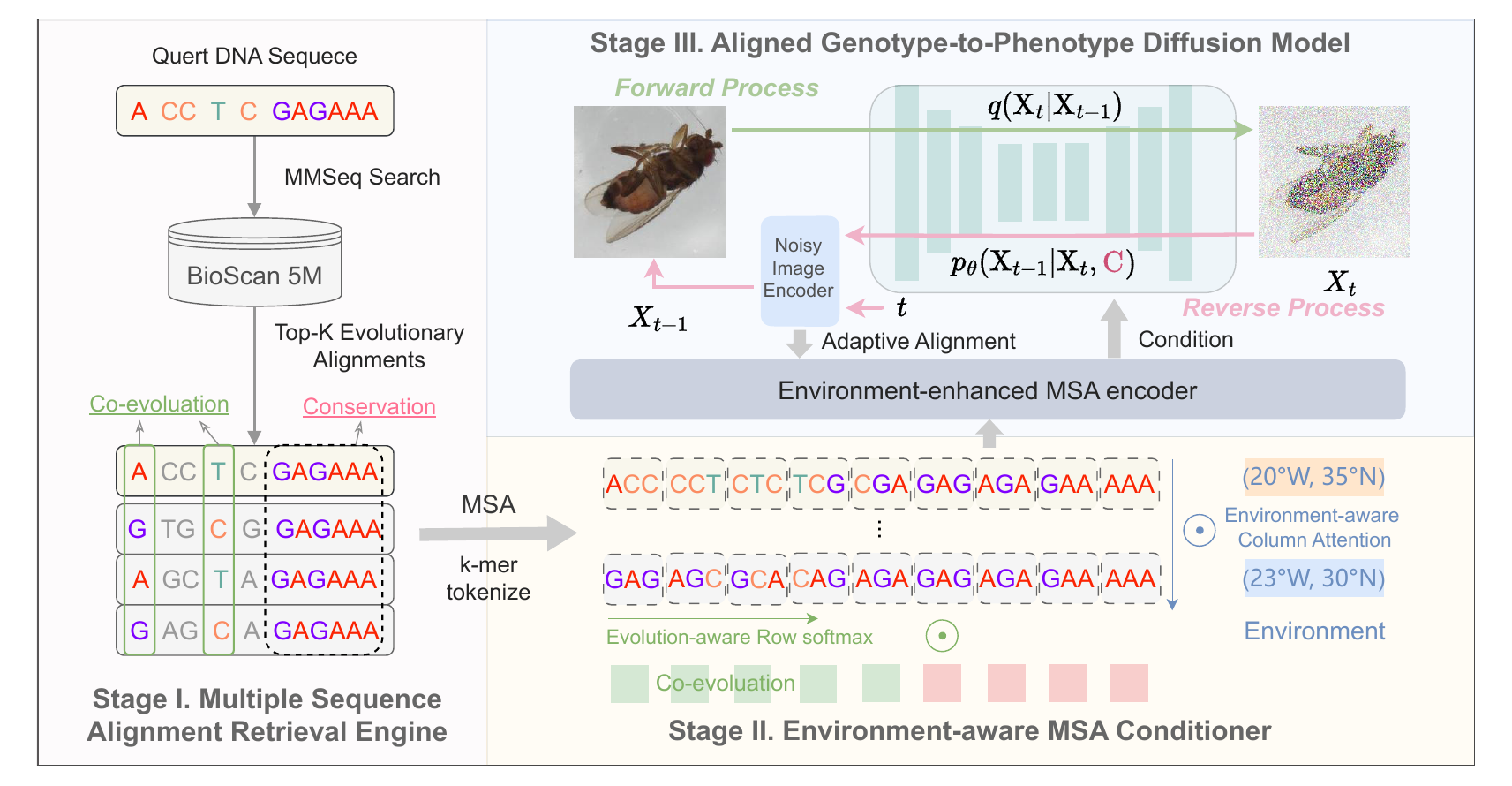}
    \vspace{-1.6em}
    \caption{\textbf{G2PDiffusion for genotype-to-phenotype image synthesis.} It first utilizes the MMseq to retrieve evolutionary alignments (in Section~\ref{sec:Constructing}). Then the retrieved MSA are fed into an environment-enhanced MSA conditioner that integrates them with environmental factors, i.e., longitude and latitude (in Section~\ref{sec:G2P}). Additionally, a cross-modality alignment guidance mechanism is employed to ensure genotype-phenotype consistency during sampling (in Section~\ref{sec:Alignment}).}
    \vspace{-1em}
    \label{fig:2_diffusion}
\end{figure*}

\paragraph{Genotype-to-Phenotype Prediction.}
Predicting phenotypes from genotypes is a fundamental challenge in biology, requiring the integration of genetic makeup and environmental influences \cite{via1985genotype,bochner2003new,dowell2010genotype}. The genotype encodes hereditary information in DNA, while the phenotype manifests as observable traits, including physical characteristics, behaviors, physiological functions, and clinical outcomes \cite{klose1999genotypes}. Here, we focus on physical characteristics as phenotypes.
Conventional genomic analysis methodologies, exemplified by genome-wide association studies (GWAS) \cite{uffelmann2021genome,tam2019benefits} and quantitative trait loci (QTL) mapping \cite{khatkar2004quantitative,kendziorski2006review}, primarily aim to identify statistical associations between genetic markers and phenotypic characteristics. The emergence of deep learning architectures—particularly convolutional neural networks (CNNs) and recurrent neural networks (RNNs)—has shifted paradigm toward decoding intricate genotype-phenotype interactions through automated pattern discovery in high-dimensional genomic datasets \cite{wang2024prediction,magnusson2022deep,danilevicz2022plant,wang2024lstm,annan2023machine}. While these computational approaches demonstrate proficiency in value regression (e.g., crop height prediction) or categorical classification (e.g., barley grain yield estimation) through supervised learning frameworks, they face notable limitations in cross-species and cross-trait generalizability due to inherent biological complexity and model dependency on domain-specific training data.  To overcome these limitations, we propose a novel paradigm that utilizes image-derived phenomic representations as biologically interpretable proxies, establishing a domain-agnostic framework for cross-species predictive modeling from morphological patterns.

\paragraph{DM-based Conditional Image Synthesis.}

Diffusion models (DMs) have shown remarkable success in generating high-quality images conditioned on additional input. Text-to-image models, such as GLIDE \cite{nichol2021glide}, Stable Diffusion \cite{rombach2022high}, and DALL-E 3 \cite{betker2023improving}, utilize semantic text encoders \cite{radford2021learning} to translate descriptive language into detailed and coherent visual outputs. Similarly, image-to-image models, including inpainting \cite{wang2023imagen,saharia2022palette,tumanyan2023plug}, super-resolution \cite{li2022srdiff,li2022srdiff}, and style transfer \cite{zhang2023inversion}, refine or transform images by leveraging diffusion-based priors.
Beyond conventional applications, DMs have been extended to specialized domains, such as medical imaging \cite{li2023zero}, where they assist in data augmentation and anomaly detection, as well as graph-to-image synthesis \cite{yang2022diffusion} and satellite imagery generation \cite{graikos2024learned}.
These advancements show the versatility of diffusion-based approaches in capturing complex structures and domain-specific patterns.

\section{Methods}
\label{sec:Preliminary}

\subsection{Problem Formulation}

We focus on the task of genotype-to-phenotype prediction, aiming to generate phenotypic images given the corresponding DNA sequence and environmental factors. Formally, the training set is denoted as $\mathcal{S} = \{(E_i, G_i, X_i)\}_{i=1}^{N_\mathcal{S}}$, where $E_i \in \mathbb{R}^2$ represents the environmental factors (e.g., longitude and latitude), $G_i \in \mathcal{G}$ denotes the DNA sequence, and $X_i \in \mathcal{X}$ represents the phenotypic image associated with $(E_i, G_i)$.
This objective is to learn a conditional generator with learnable parameters $\theta$:
\begin{align}
\label{eq:forward_ddpm2}
    f_{\theta}: (E, G) \rightarrow X.
\end{align}


\subsection{Framework Overview}
\label{sec:Framework Overview}


We propose G2PDiffusion, a novel evolution-aware diffusion framework for genotype-to-phenotype image synthesis, as shown in Figure~\ref{fig:2_diffusion}.
It contains three novel components: a highly-efficient MSA retrieval engine, an environment-aware MSA conditioner, and a dynamic phenomic alignment module.
Firstly, the MSA engine retrieves MSA from an external database to identify evolutionarily conserved and variable sequence regions (Section~\ref{sec:Constructing}). Then, the retrieved MSA together with environment contexts are fed into a conditional encoder to learn the genotype-environment (GxE) interaction (Section~\ref{sec:G2P}). Finally, we build a diffusion model based on GxE representation to generate images recording morphological features. 
During each denoising step, a dynamic phenomic alignment module is employed to refine phenotypic representations
(Section~\ref{sec:Alignment}).

\subsection{M\hspace{-0.15pt}u\hspace{-0.15pt}l\hspace{-0.15pt}t\hspace{-0.15pt}i\hspace{-0.15pt}p\hspace{-0.15pt}l\hspace{-0.15pt}e \hspace{-0.15pt}S\hspace{-0.15pt}e\hspace{-0.15pt}q\hspace{-0.15pt}u\hspace{-0.15pt}e\hspace{-0.15pt}n\hspace{-0.15pt}c\hspace{-0.15pt}e \hspace{-0.15pt}A\hspace{-0.15pt}l\hspace{-0.15pt}i\hspace{-0.15pt}g\hspace{-0.15pt}n\hspace{-0.15pt}m\hspace{-0.15pt}e\hspace{-0.15pt}n\hspace{-0.15pt}t\hspace{-0.15pt}s \hspace{-0.15pt}R\hspace{-0.15pt}e\hspace{-0.15pt}t\hspace{-0.15pt}r\hspace{-0.15pt}i\hspace{-0.15pt}e\hspace{-0.15pt}v\hspace{-0.15pt}a\hspace{-0.15pt}l \hspace{-0.15pt}E\hspace{-0.15pt}n\hspace{-0.15pt}g\hspace{-0.15pt}i\hspace{-0.15pt}n\hspace{-0.15pt}e}

\label{sec:Constructing}
Considering that Multiple Sequence Alignments (MSA) aggregate homologous DNA sequences across species, it serves as a crucial tool for capturing evolutionary constraints and conserved functional regions in the DNA sequence \citep{zhang2024historical, zou2015halign}. We utilize MMseqs2 \citep{steinegger2017mmseqs2}, a fast and scalable sequence search tool, to construct evolutionary alignments by retrieving homologous sequences from a reference database.
Given a query DNA sequence $G_q$ and an external sequence database $\{G_i\}_{i=1}^{N_{\mathcal{D}}}$, we utilize MMseqs2’s sensitive search module to efficiently scan the database and retrieve a set of homologous sequences with top-$m$ high evolutionary similarity.
We write the retrieved homologous sequence pool:
\begin{equation}
\mathcal{D}(G_q,m) := \text{top}_m (\{G_i\}_{i=1}^{N_{\mathcal{D}}}, \text{MMseqs}(\cdot, G_q)).
\end{equation}


The resulting MSA captures conserved sequence motifs, co-evolutionary relationships, and functionally significant variations, providing a biologically meaningful prior for guiding the phenotype synthesis process.



\subsection{Environment-aware MSA Conditioner}
\label{sec:G2P}
To accurately capture the genotype-to-phenotype mapping across diverse species, it is necessary to design a conditioner that captures the complex interplay between MSA-derived genetic information and environmental contexts.
The MSA-derived genetic information reveals how conserved regions maintain core functions and variable sites contribute to phenotypic diversity, while environmental factors drive phenotypic adaptation through selective pressures over evolutionary timescales. Methodological details are as follows:






\paragraph{$k$-mer Tokenization \& Input Format.}
DNA sequences, consisting of long chains of nucleotides (adenine, cytosine, guanine, and thymine), are inherently complex and require a systematic approach to capture meaningful patterns.
Instead of regarding each base as a single token, we tokenize a DNA sequence with the $k$-mer representation \cite{chor2009genomic}, an approach that has been widely used in analyzing DNA sequences.
This method treats a subsequence of $k$ consecutive nucleotides as a ``word" to be tokenized, enabling the model to capture local sequence patterns and motifs that may influence phenotypic traits. For example, given the original DNA sequence $G=[A,C,C,T,C,...]$, the $3$-mer representation is $\texttt{Tokenizer}(G,3)=[ACC, CCT,CTC,...]$. 
We write the $k$-mer token as $\mathcal{T}=\texttt{Tokenizer}(G,k)=[t_1, t_2, ..., t_l]$, where $l$ is the length of the tokenized sequence. When retrieving $m$ MSA sequences $\{G_i\}_{i=1}^m$, we compare nucleotides column-wise to obtain the evolution vector $V = [v_1, v_2, \dots, v_l]$, where $v_i \in \{0,1\}$ indicates whether position $i$ is conserved: if all nucleotides in the column are identical, $v_i = 1$; otherwise, $v_i = 0$. The evolution vector, tokenized MSA, and environments $\{E_i\}_{i=1}^m$ together form the complete input as:
\begin{equation*}
\langle
    \begin{array}
    {c@{\, ,\,}c@{\, ,\,}c}
        \begin{bmatrix}
            v_1 & v_2 & \cdots & v_l
        \end{bmatrix} 
        &
        \begin{bmatrix}
            \mathcal{T}_{1,1} & \mathcal{T}_{1,2} & \cdots  &
            \mathcal{T}_{1,l}
            \\
            \mathcal{T}_{2,1} & \mathcal{T}_{2,2} & \cdots  &
            \mathcal{T}_{2,l}
            \\
            \cdots & \cdots & \cdots & \cdots 
            \\
            
            \mathcal{T}_{m,1} & \mathcal{T}_{m,2} & \cdots  &
            \mathcal{T}_{m,l}
            
        \end{bmatrix} 
        &
        \begin{bmatrix}
            E_{1} \\
            E_{2} \\
            \cdots  \\
            E_{m} 
        \end{bmatrix}
        
    \end{array}
\rangle
\end{equation*}
where $v_i \in \{0,1\}$ indicates whether position $i$ is conservation, $\mathcal{T}_{i,j}$ represents the $j$-th token in the $i$-th MSA token sequence, and $E_i$ is the $i$-th environment.




\paragraph{Evolution-aware Row Attention.} 
We employ an MLP to transform the evolution vector into a gating weight \(\boldsymbol{w}_v \in \mathbb{R}^{1 \times l}\), which modulates the row attention mechanism:  
\begin{equation}
    H^{\text{row}}_{i,:} = \text{Softmax} \left( \frac{Q(\mathcal{T}_{i,:}) K(\mathcal{T}_{i,:})^\top \odot \boldsymbol{w}_v}{\sqrt{d}} \right) V(\mathcal{T}_{i,:}),
\end{equation}
where \( Q(\cdot), K(\cdot), V(\cdot) \) are MLPs for computing query, key, and value, respectively; $\odot$ denotes element-wise multiplication, \( d \) is the feature dimension, and \(\mathcal{T}_{i,:}\) is the \(i\)-th row of the MSA matrix.
Row attention applies position-wise evolutionary gating to the intra-sequence attention weights, allowing adaptive modulation of attention scores for conserved and evolutionary regions.



\paragraph{Environment-aware Column Attention.} Considering the inherent spherical nature of Earth's geographic coordinates, we map latitude $(\beta)$ and longitude $(\lambda)$ to spherical coordinates $(x,y,z)=(\cos\beta \cos\lambda, \cos\beta  \sin\lambda, \sin\beta )$, which are fed to an MLP to obtain the environment weighting vector $\boldsymbol{w}_e \in \mathbb{R}^{m \times 1} $. We take $\boldsymbol{w}_e$ to modulate the column attention:
\begin{equation}
    H^{\text{col}}_{:,j} = \text{Softmax} \left( \frac{Q(\mathcal{T}_{:,j}) K(\mathcal{T}_{:,j})^\top \odot \boldsymbol{w}_e}{\sqrt{d}} \right) V(\mathcal{T}_{:,j}),
\end{equation}
where \(\mathcal{T}_{:,j}\) represents the \(j\)-th row of the MSA matrix. The column attention mechanism injects environment information into cross-sequence representation learning.




\paragraph{Environment-enhanced MSA Encoder.}
To incorporate both genetic and environmental information, we employ a transformer-based encoder using row and column attentions that integrate information from multiple sequence alignments (MSAs) and environmental features. In addition, we use the MSA LayerNorm to stabilize model training: 
\begin{equation}
    H^{\text{MSA}} = \text{LayerNorm} \left( H^{\text{row}} + H^{\text{col}} \right).
\end{equation}
In the final layer, we obtain the genotype-environment representation via average pooling:
\begin{align}
\label{eq:condition}
C = \frac{1}{m} \sum_{i=1}^{m} H^{\text{MSA}}_{i,j} \in \mathbb{R}^{l\times d},
\end{align}
where $m$ represents the number of sequences in MSA.
This conditioning strategy enables the generative model to leverage MSA evolution patterns and environmental dependencies, leading to biologically plausible phenotype synthesis.

\subsection{A\hspace{-0.26pt}l\hspace{-0.26pt}i\hspace{-0.26pt}g\hspace{-0.26pt}n\hspace{-0.26pt}e\hspace{-0.26pt}d \hspace{-0.4pt}G\hspace{-0.26pt}e\hspace{-0.26pt}n\hspace{-0.26pt}o\hspace{-0.26pt}t\hspace{-0.26pt}y\hspace{-0.26pt}p\hspace{-0.26pt}e-t\hspace{-0.26pt}o-P\hspace{-0.26pt}h\hspace{-0.26pt}e\hspace{-0.26pt}n\hspace{-0.26pt}o\hspace{-0.26pt}t\hspace{-0.26pt}y\hspace{-0.26pt}p\hspace{-0.26pt}e\hspace{-0.4pt} D\hspace{-0.26pt}i\hspace{-0.26pt}f\hspace{-0.26pt}f\hspace{-0.26pt}u\hspace{-0.26pt}s\hspace{-0.26pt}i\hspace{-0.26pt}o\hspace{-0.26pt}n\hspace{-0.4pt} M\hspace{-0.26pt}o\hspace{-0.26pt}d\hspace{-0.26pt}e\hspace{-0.26pt}l}

\label{sec:Alignment}

We propose an aligned genotype-to-phenotype diffusion model, which leverages a conditional diffusion backbone enhanced by a dynamic cross-modality alignment mechanism to improve the consistency between generated phenotypic images and the corresponding genotypic information.

\vspace{-3.5mm}
\paragraph{Conditional Genotype-to-Phenotype Diffusion Models.}
\label{sec:diffusion}
Inspired by the success of conditional diffusion models in text-to-image generation \cite{ho2020denoising, betker2023improving,saharia2022photorealistic,zhang2023adding}, we adopt a conditional diffusion framework where the condition is the learned GxE representation $C$. The diffusion process consists of two main stages: forward diffusion and reverse denoising\cite{ho2020denoising}.

During the forward process, gaussian noise is progressively added to a phenotypic image $X_0$ over $T$ steps, which is formally defined as a Markov chain:
\begin{align}
    q\left(X_{t} \mid X_{t-1}\right)=\mathcal{N}\left(X_{t} \mid \sqrt{\alpha_{t}} X_{t-1},\left(1-\alpha_{t}\right) \bf I\right).
\end{align}
Here, $\alpha_{t}$ controls the noise intensity. By denoting $\Bar{\alpha}_t = \prod\nolimits_{i=1}^{t} \alpha_i$, we can describe the entire diffusion process as:
\begin{gather}
    q\left( X_{1:T} \mid X_{0} \right) = \prod\nolimits_{t=1}^{T} q \left( X_t \mid X_{t-1} \right), \\
    q \left( X_t \mid X_{0} \right) = \mathcal{N} \left( X_t; \sqrt{\Bar{\alpha}_t} X_{0}, (1 - \Bar{\alpha}_t) \bf I \right).
\end{gather}


During the reverse process, it gradually removes noise from the sample $X_T$, eventually recovering $X_0$. A denoising model $\epsilon_{\theta} (X_t, t, C)$ is trained to estimate the noise $\epsilon$ from $X_t$ and a condition embedding $C$, 
which is formally denoted as 
\begin{align}
    p_{\theta} \left( X_{t-1} | X_{t}, t, C \right) &= \mathcal{N} \left( X_{t-1}; \epsilon_{\theta} (X_t, t, C), \sigma_{t}^{2} \bf I \right).
\end{align}

The denoising process is trained by maximizing the likelihood of the data under the model or, equivalently, by minimizing the variational lower bound on the negative log-likelihood of the data. \cite{ho2020denoising} shows that this is equivalent to minimizing the KL divergence between the predicted distribution $p_{\theta} (X_{t-1} | X_{t}, C)$ and the ground-truth distribution $q (X_{t-1} | X_{t}, X_{0}, C)$ at each time step $t$ during the backward process. The training objective then becomes:
\begin{align}
    \min\nolimits D_{KL} \left( q \left( X_{t-1} | X_{t}, X_{0}, C \right) \big\| p_{\theta} \left( X_{t-1} | X_{t}, C \right) \right),
\end{align}
which can be simplified as:
\begin{align}
    L_{DM} &= \mathbb{E}_{\epsilon, t} \left[\| \epsilon - \epsilon_{\theta} (X_t, t, C) \|_{2}^{2}\right]. 
    \label{eq:loss}
\end{align}

\paragraph{Dynamic Alignment Sampling Mechanism.}    To enhance genotype-phenotype consistency, we introduce a cross-modal alignment strategy that integrates the reverse diffusion process with the MSA encoder. 
Specifically, we propose a gradient-guided alignment framework, where an alignment model $g_{\phi}(X_t, t)$ is trained to align noisy image embedding $X_t$ to the associated DNA embedding.
This process, termed dynamic alignment, leverages noisy images at multiple diffusion steps to refine phenotype representations.
Mathematically, the conditional diffusion score \cite{ho2022classifier} is 
{\small
\begin{align*}
    \epsilon(X_t, t, C) \approx -\sqrt{1-\alpha_t} \nabla_{X_t} \left[\log{p_{\theta}(X_t|C)} + w \log{p_{\phi}(C|X_t)}\right],
\end{align*}
}
where $w$ controls the strength of alignment guidance.
We define the learning objective of the aligner $g_{\phi}(\cdot,\cdot)$  as
\begin{align}
\label{eq:aligner}
    \gL_{align} = - \log \frac{\exp{\left[g_{\phi}(X_t,t) \cdot C^+ \right]}}{\sum_{j=1}^B \exp{\left[g_{\phi}(X_t,t) \cdot C_j \right]}},
\end{align}
where $\phi$ is learnable parameter, $B$ is  batch size, $X_t$ is the noised image at diffusion step $t$,  $C^+$ is the ground-truth GxE representation related to the phenotype. 

\vspace{-3mm}
\paragraph{Sampling.} Compared to previous research \cite{kim2022diffusionclip} that directly uses CLIP loss for gradient guidance, our method can dynamically align noisy images to the DNA embeddings during diffusion trajectory, which is better suited to the noisy nature of the diffusion process \cite{nichol2021glide}.
Algorithm~\ref{alg:sample} summarizes the guided genotype-to-phenotype sampling process.

\begin{algorithm}
\caption{Diffusion Model Sampling with Guidance}
\label{alg:sample}
\begin{algorithmic}[1]
\STATE \textbf{Input:} Initial noise $X_T$, DNA sequence $G_q$, environment context $E$,  retrival database $\mathcal{D}$, environment-enhanced MSA encoder $\mathcal{C}(\cdot)$, conditional diffusion model $\epsilon_{\theta}(X_t, t, C)$, aligner $g_{\phi}(X_t, t)$, guidance strength $w$, update rate $\eta$

\STATE \textbf{Initialize} $X_T$ as random noise
\STATE Retrieve $m$ similar DNA sequences $\{G_i\}_{i=1}^m$ from $\mathcal{D}$ according to $G_q$, and get GxE representation $C= \mathcal{C}(G_q, \{G_i\}_{i=1}^m, E)$
\FOR{$t = T$ down to 1}
    \STATE Compute $\nabla_{X_t} \log p_{\theta}(X_t| C)$ using the conditional diffusion model $\epsilon_{\theta}(X_t, t, C)$;  
    \STATE Compute $\gL_{align}$ using the aligner $g_{\phi}(X_t, t)$ and $C$, referring to Eq.~\ref{eq:aligner};
    \STATE Update gradient:\\
    \scalebox{0.87}{ $\nabla_{X_t}\log p_{\theta,\phi}(X_t| C) \leftarrow \nabla_{X_t}\log p_{\theta}(X_t|C)+w\nabla_{X_t} \gL_{align}$}
    \STATE Estimate $X_{t-1}$ using the updated gradient:\\
    $
    X_{t-1} = X_{t} - \eta \cdot \nabla_{X_{t}} \log p_{\theta, \phi}(X_{t} \mid C)
    $
    \ENDFOR
\STATE \textbf{Output:} Sample $X_0$
\end{algorithmic}
\end{algorithm}

\vspace{-3mm}
\section{Evaluation Setup}

\paragraph{Dataset.}
We used the BIOSCAN-5M dataset \cite{gharaee2024bioscan5m}, the largest multi-modal resource available for genotype-to-phenotype prediction.
It contains over 5 million insect specimens with taxonomic labels, DNA barcode sequences, geographic coordinates (longitude and latitude), and phenotypic images. We preprocessed the phenotypic images by resizing and padding them to a resolution of 
$256 \times 256$. The seen-set images were then split into training and validation sets using a 90-10 ratio. Additionally, the dataset includes an unseen set, consisting of samples either lacking species labels or belonging to organisms without established scientific names.


\vspace{-3mm}
\paragraph{Baselines.}
Since no direct baselines for genotype-to-phenotype image synthesis, we employ a comparative framework that adapts the leading conditional image generation methods to this specialized task. The baselines include GAN-based approaches such as DF-GAN \cite{liao2022text},  diffusion-based methods like Stable Diffusion \cite{rombach2022high}, and ControlNet \cite{zhang2023adding}.


\vspace{2mm}
\textit{We introduce the following new metrics for the genotype-to-phenotype prediction task}:
\vspace{-3mm}

\paragraph{CLIBDScore.}
This metric is built on the pre-trained CLIBD model \cite{gong2024clibd} to measure the semantic similarity between the DNA and image, which uses CLIP-style \cite{radford2021learning} contrastive learning to align images and barcode DNA representations in a unified embedding space.
Similar to CLIPScore \cite{hessel2021clipscore}, a commonly-used metric for text-image alignment, \textbf{CLIBDScore} measures how well an image-based morphology is aligned with the corresponding DNA.

\begin{figure}[h]
    \centering
    \includegraphics[width=0.9\linewidth]{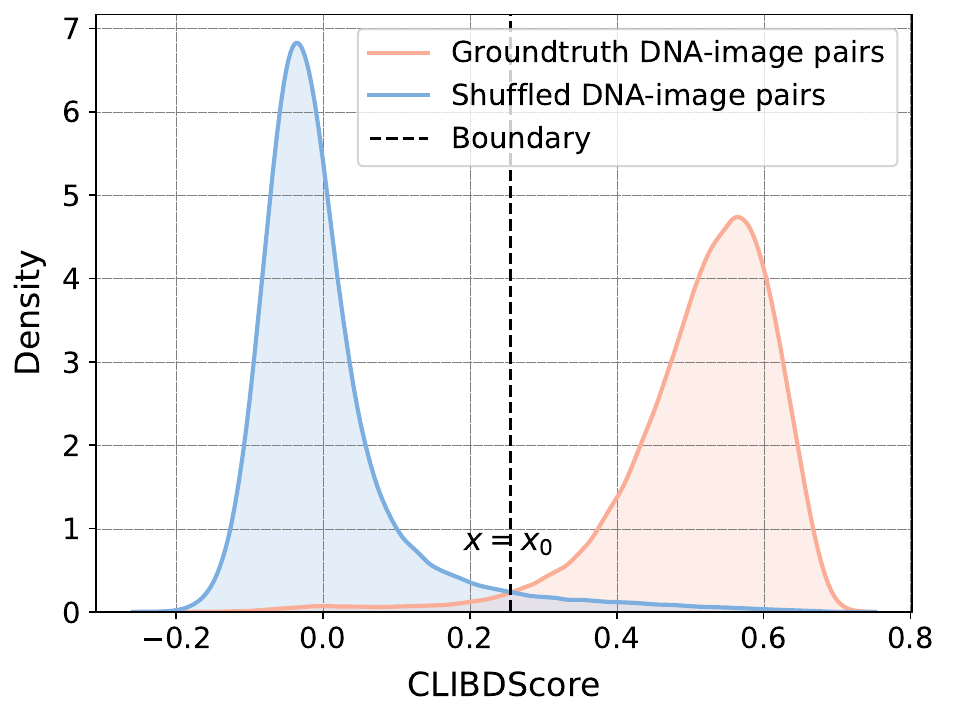}
    \vspace{-2mm}
    \caption{Density Distribution of DNA-Image CLIBDScore.}
    \label{fig:CLIBDScore_Distribution}
    \vspace{-8mm}
\end{figure}

\paragraph{Success Rate.}
Moreover, we compute the CLIBDScore for all DNA-image pairs in the training set, and the randomly shuffled pairs for comparison. 
The density distributions of these two sets are illustrated in Fig.~\ref{fig:CLIBDScore_Distribution}.
The minimal overlap between the two distributions indicates that true DNA-image pairs are distinguishable from the random pairs. 
Building on this observation, we introduce an additional evaluation metric, \textbf{Success Rate}, which is based on the intersection line $x=x_0=0.255$ between the two distributions. If CLIBDScore exceeds this threshold $x_0$, the prediction is considered successful; otherwise, it is considered a failure.

\vspace{-5mm}
\paragraph{PES.} We introduce Phenotype Embedding Similarity (PES) as a metric to assess the biological relevance of generated images by comparing them to real images in a learned phenotype feature space. Specifically, we first train a species classification model using authentic phenotype images, with an intermediate embedding layer that captures species-specific visual characteristics. During evaluation, both real and generated images are processed through the classifier to extract their embeddings, and PES is calculated as the average cosine similarity between the embeddings of real and generated images corresponding to the same DNA. Higher PES values indicate that the generated images more accurately preserve species-level phenotypic features, providing a biologically meaningful measure of image quality.



\vspace{-5mm}
\paragraph{Implementation details.} All the models are trained on 8 NVIDIA-A100 GPUs using Adam optimizer \cite{kingma2014adam} up to 100k steps, with the learning rate of 1e-5, batch size of 128 and cosine annealing scheduler. 
During sampling, for each given DNA, we generate $n$ images, compute CLIBDScore, Success Rate, and PES metrics, and record the hignest score as the top-$n$ values ($n$ takes values of 1, 5, 10, 20, 50, and 100, as shown in the following experimental sections).





\section{Results}
\label{sec:experiments}
In this section, we conduct extensive experiments to answer the following questions: 
\begin{itemize}[leftmargin=5.5mm]

  \item \textbf{Performance (Q1):} 
  Could the model generate phenotypic images that match the DNA?
  \item \textbf{Model Analysis (Q2):} What is the impact of each module on the model's overall performance?
  \item \textbf{Generalization (Q3):} Could our proposed method generalize across unseen species?
\end{itemize}

\begin{table*}[t]
\centering
\caption{Summary of CLIBDScore and success rate at different thresholds of ground-truth images, as well as images generated by our model and other non-diffusion and diffusion-based baselines. Our method outperforms the baselines across all evaluation metrics.}
\label{table:Q1}
\begin{adjustbox}{width=1\linewidth}
    \begin{tabular}{c|l|c|c|cc|cc|cc|cc}
    \toprule
    \textbf{Metric} & \textbf{Rank} & \textbf{GT} & \textbf{Random} &  \multicolumn{2}{c|}{\textbf{DF-GAN}} & \multicolumn{2}{c|}{\textbf{Stable Diffusion}} & \multicolumn{2}{c|}{\textbf{ControlNet}} &  \multicolumn{2}{c}{\chl \textbf{G2PDiffusion}} \\
     & & & & Abs. & Rel. & Abs. & Rel. & Abs. & Rel. & \chl Abs. & \chl Rel. \\
    \toprule
    \multirow{6}{*}{CLIBDScore} & Top-1 & \multirow{6}{*}{0.512} & \multirow{6}{*}{0.005} & 0.054 & 0.106 & 0.100 & 0.195 & 0.107 & 0.209 & \chl \textbf{0.182} & \chl \textbf{0.356} \\
    & Top-5 &&& 0.154 & 0.301 & 0.219 & 0.428 & 0.228 & 0.445 & \chl \textbf{0.302} &  \chl \textbf{0.590} \\
    & Top-10 &&& 0.181 & 0.354 & 0.254 & 0.496 & 0.265 & 0.518 & \chl \textbf{0.358} & \chl \textbf{0.700} \\
    & Top-20 &&& 0.224 & 0.438 & 0.292 & 0.570 & 0.307 & 0.600 & \chl \textbf{0.397} & \chl \textbf{0.776} \\
    & Top-50 &&& 0.276 & 0.539 & 0.338 & 0.660 & 0.351 & 0.686 & \chl \textbf{0.455} & \chl \textbf{0.889} \\
    & Top-100 &&& 0.314 & 0.614 & 0.367 & 0.718 & 0.384 & 0.750 &\chl  \textbf{0.480} & \chl \textbf{0.938} \\
    \toprule
    \multirow{6}{*}{Success Rate} & Top-1 & \multirow{6}{*}{96.4\%} & \multirow{6}{*}{4.4\%} & 5.6\% & 5.8\% & 11.5\% & 11.9\% & 12.4\% & 12.9\% &\chl \textbf{31.7\%} & \chl \textbf{32.8\%} \\
    & Top-5 &&& 18.7\% & 19.4\% & 36.6\% & 38.0\% & 39.1\% & 40.6\% & \chl \textbf{65.8\%} & \chl \textbf{68.3\%} \\
    & Top-10 &&& 32.1\% & 33.3\% & 43.5\% & 45.1\% & 47.0\% & 48.7\% & \chl \textbf{81.1\%} & \chl \textbf{84.1\%} \\
    & Top-20 &&& 40.9\% & 42.4\% & 55.7\% & 57.7\% & 57.8\% & 60.0\% & \chl \textbf{90.4\%} & \chl \textbf{93.8\%} \\
    & Top-50 &&& 48.1\% & 49.9\% & 68.7\% & 71.3\% & 70.7\% & 73.4\% & \chl \textbf{93.0\%} & \chl \textbf{96.5\%} \\
    & Top-100 &&& 52.6\% & 54.6\% & 74.8\% & 77.6\% & 77.0\% & 79.8\% & \chl \textbf{94.0\%} & \chl \textbf{97.5\%} \\
    
    \bottomrule
    \end{tabular}
\end{adjustbox}
\vspace{-3mm}
\end{table*}

\vspace{-1mm}

\subsection{Performance (Q1)}

\begin{figure}[h]
    \centering
    \vspace{-1em}
    \includegraphics[width=1\linewidth]{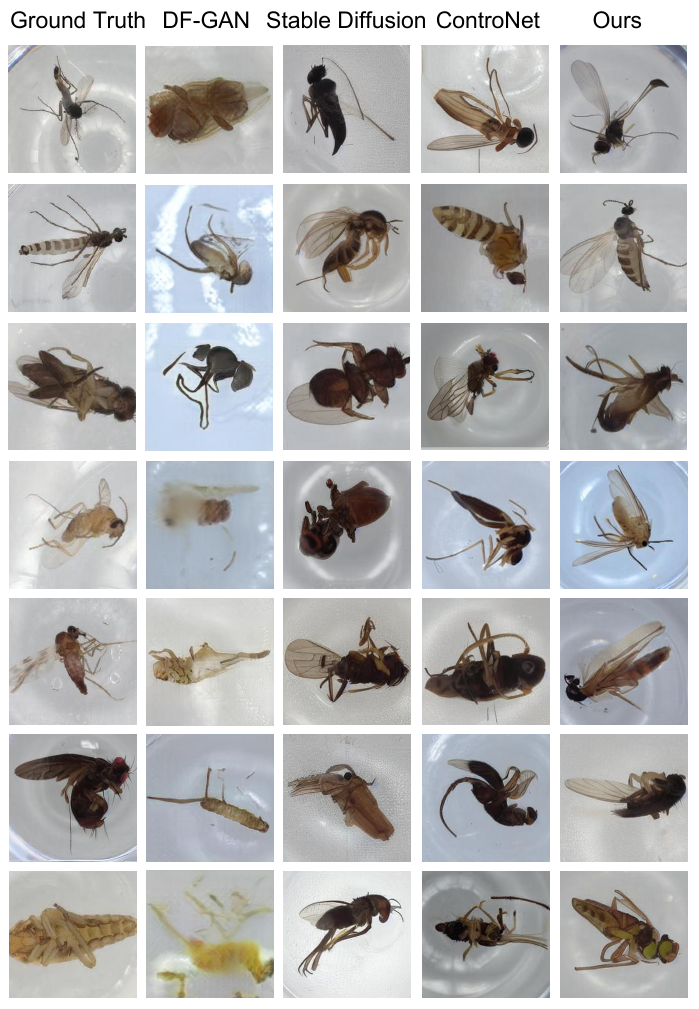}
    \vspace{-1.5em}
    \caption{Generative results. All methods can generate visually reasonable images with different the DNA-image consistency.}
    \vspace{-2em}
    \label{fig:Q1_Qualitative_results}
\end{figure}

\paragraph{Qualitative Results.} 
Fig.~\ref{fig:Q1_Qualitative_results} shows the qualitative results of various methods. Our method, G2PDiffusion, stands out by producing the most resonable phenotype predictions from DNA inputs, thanks to the carefully designed evolutionary conditioner and dynamic aligner. DF-GAN, on the other hand, struggles to generate high-quality images and often fails to capture the precise characteristics of the ground truth phenotypes. Although Stable Diffusion and ControlNet could generate visually appealing images, they lack the ability to align these images closely with the true phenotypes.

\begin{table}[H]
\centering
\caption{PES scores comparison across DF-GAN, Stable Diffusion, ControlNet, and our proposed G2PDiffusion. G2PDiffusion achieves the highest PES scores at all evaluated ranks.}
\label{tab:pes_comparison}
\begin{adjustbox}{width=1\linewidth}
    \begin{tabular}{l|c|c|c|c}
    \toprule
    Rank & DF-GAN & Stable Diffusion & ControlNet &  \chl G2PDiffusion \\
    \midrule
    Top-1   & 0.021 & 0.062 & 0.061 & \chl \textbf{0.152} \\
    Top-5   & 0.134 & 0.207 & 0.212 & \chl \textbf{0.291} \\
    Top-10  & 0.167 & 0.240 & 0.254 & \chl \textbf{0.346} \\
    Top-20  & 0.216 & 0.288 & 0.299 &\chl  \textbf{0.405} \\
    Top-50  & 0.276 & 0.349 & 0.359 & \chl \textbf{0.478} \\
    Top-100 & 0.301 & 0.389 & 0.403 & \chl \textbf{0.511} \\
    \bottomrule
    \end{tabular}
\end{adjustbox}
\end{table}

\paragraph{Quantitative results.}
For quantitative evaluation, we consider the three metics: CLIBDScore, Success Rate, and Phenotype Embedding Similarity (as shown in Table \ref{table:Q1} and Table \ref{tab:pes_comparison}).
In addition to reporting absolute scores, we also calculate relative scores by dividing each score by the ground truth score (shown as Abs. and Rel. in the table). We summary that:
(a) Compared to the random baseline, all deep learning methods demonstrate non-trivial potential in deciphering phenotypes from genotype and environment. (b) Diffusion models consistently outperform DF-GAN, as their multi-step generation process progressively refines the generated phenotypes, making it easier to capture the complex genotype-phenotype relationships. (c) The proposed G2PDiffusion demonstrates significantly higher performance than other models across all metrics. For example, in the Top-5 success rate, our model achieves a score of 65.8\%, notably outperforming Stable Diffusion (36.6\%) and ControlNet (39.1\%). Furthermore, our method shows remarkable improvements with a Top-10 success rate of 81.1\% and a Top-100 rate of 94.0\%, indicating strong alignment with ground truth images. These results highlight the effectiveness of our approach in accurately generating phenotype images from DNA sequences. (d) The compared PES scores show that G2PDiffusion generates morphological phenotypes with higher biological relevance in the phenotype embedding space, demonstrating that incorporating genotype-environment interaction and evolutionary constraints helps align generated images with real phenotypic variation.

\begin{table*}[t]
\caption{Summary of CLIBDScore and success rate evalutions at different thresholds on the unseen set. }
\label{table:Q2_unseen_compare}
\centering
{\resizebox{\textwidth}{!}{
\begin{tabular}{l|cc|cc|cc|cc|cc|cc}
\toprule
\textbf{Method} & \multicolumn{2}{c|}{\textbf{Top-1}} & \multicolumn{2}{c|}{\textbf{Top-5}} & \multicolumn{2}{c|}{\textbf{Top-10}} & \multicolumn{2}{c|}{\textbf{Top-20}}   & \multicolumn{2}{c|}{\textbf{Top-50}}  & \multicolumn{2}{c}{\textbf{Top-100}} \\
& Score. & Acc. & Score. & Acc. & Score. & Acc. & Score. & Acc. & Score. & Acc. & Score. & Acc.\\
\toprule
DF-GAN & 0.045 & 4.2\% & 0.110 & 12.5\% & 0.130 & 18.3\% & 0.155 & 22.8\% & 0.180 & 33.7\% & 0.190 & 38.4\% \\
Stable Diffusion & 0.068 & 6.4\% & 0.162 & 19.3\% & 0.185 & 28.7\% & 0.210 & 37.5\% & 0.235 & 48.2\% & 0.250 & 53.1\% \\
ControlNet & 0.072 & 7.1\% & 0.155 & 18.4\% & 0.180 & 29.2\% & 0.205 & 40.3\% & 0.235 & 51.7\% & 0.250 & 56.3\% \\
\chl Ours &\chl  \textbf{0.081} &\chl  \textbf{8.8\%} &\chl  \textbf{0.184} &\chl  \textbf{25.0\%} &\chl  \textbf{0.228} &\chl  \textbf{41.4\%} &\chl  \textbf{0.263} &\chl  \textbf{55.1\%} &\chl  \textbf{0.313} &\chl  \textbf{75.5\%} &\chl  \textbf{0.340} &\chl  \textbf{80.3\%} \\
\bottomrule
\end{tabular}
}}
\vspace{-1em}
\end{table*}

\subsection{Model Analysis (Q2)}
\paragraph{Effects of Environment-aware MSA Conditioner and Dynamic Alignment.}
We investigate the impact of environment-aware MSA conditioner and dynamic alignment sampling mechanism, as shown in Table \ref{tab:ablation}. In particular, we replace the environment-aware MSA encoder with the simplest DNABERT\cite{ji2021dnabert} and remove the dynamic alignment sampling mechanism to construct our baseline.

\begin{table}[h]
\centering
\caption{Ablation studies of environment-aware MSA conditioner and dynamic alignment sampling mechanism.}
\label{tab:ablation}
\begin{adjustbox}{width=1\linewidth}
    \begin{tabular}{l|cc|cc|cc}
    \toprule
    \multicolumn{1}{c|}{Methood} & \multicolumn{2}{c|}{CLIBDScore} & \multicolumn{2}{c|}{Success Rate} & \multicolumn{2}{c}{PES} \\
    & Top-1 & Top-5 & Top-1 & Top-5 & Top-1 & Top-5 \\
    \midrule
    Baseline & 0.100 & 0.219 & 11.50\% & 26.60\% & 0.062 & 0.187 \\
    + Conditioner & 0.125 & 0.235 & 16.73\% & 28.21\% & 0.098 & 0.254 \\
    + Alignment & 0.167 & 0.289 & 27.14\% & 51.24\% & 0.137 & 0.268 \\
    + Both & 0.182 & 0.302 & 31.70\% & 65.80\% & 0.152 & 0.291 \\
    \bottomrule
    \end{tabular}
\end{adjustbox}
\vspace{-3mm}
\end{table}

The ablation results show that both the environment-aware MSA conditioner and the dynamic alignment sampling mechanism contribute to model performance. We summary that incorporating evolutionary context and environment-aware sequence representations helps the model capture biologically meaningful genotype-phenotype relationships. Meanwhile, the dynamic alignment sampling mechanism further enhances the biological relevance of generated phenotypes to the DNA sequences.

\paragraph{Effects of Evolutional-Alignments Retrieval}

We investigate the influence of the retrieved MSA for G2PDiffusion through an ablation study on variable $m$, which denotes the number of retrieved sequence alignments.
From the results in Table \ref{tab:ablation_m}, we observe that: 
(a) increasing $m$ from 0 to 1 leads to significant improvements across all evaluation metrics, indicating that incorporating homologous sequence alignments provides evolutionary context, which enhances the quality of phenotype generation; (b) the best performance is achieved when $m$ is set to 1 or 2, where the retrieved sequences exhibit high similarity to the target, enabling effective integration of conserved evolutionary signals into the generation process; (c) however, further increasing $m$ introduces more distant sequences with lower relevance, which inevitably introduces noise and reduces the overall generation quality.

\begin{table}[h]
\centering
\caption{The effect of hyper-parameter $m$. The top 2 results are highlighted with \textbf{bold text} and \underline{underlined text}, respectively.}
\label{tab:ablation_m}
\begin{adjustbox}{width=1\linewidth}
    \begin{tabular}{l|cc|cc|cc}
    \toprule
    \multicolumn{1}{c|}{Methood} & \multicolumn{2}{c|}{CLIBDScore} & \multicolumn{2}{c|}{Success Rate} & \multicolumn{2}{c}{PES} \\
    & Top-1 & Top-5 & Top-1 & Top-5 & Top-1 & Top-5 \\
    \midrule
    $m$=0 & 0.178 & 0.284 & 27.17\% & 53.10\% & 0.151 & 0.284 \\
    $m$=1 & \textbf{0.193} & \underline{0.299} & \textbf{36.23\%} & \textbf{65.80\%} & \underline{0.143} & \textbf{0.293} \\
    $m$=2 & \underline{0.182} & \textbf{0.302} & \underline{31.70}\% & \textbf{65.80\%} & \textbf{0.152} & \underline{0.291} \\
    $m$=3 & 0.166 & 0.285 & 29.90\% & 58.87\% & 0.128 & 0.271 \\
    $m$=4 & 0.176 & 0.296 & 29.40\% & 62.34\% & 0.142 & 0.280 \\
    \bottomrule
    \end{tabular}
\end{adjustbox}
\end{table}

\subsection{Generalization to Unseen Species (Q3)}


To investigate the generalization capability of our method, we evaluate its performance on unseen species in the dataset, called the \textbf{open-world scenario}. In this case, species do not have scientific names in the dataset.

\begin{figure}[!htb]
    \centering
    \includegraphics[width=1\linewidth]{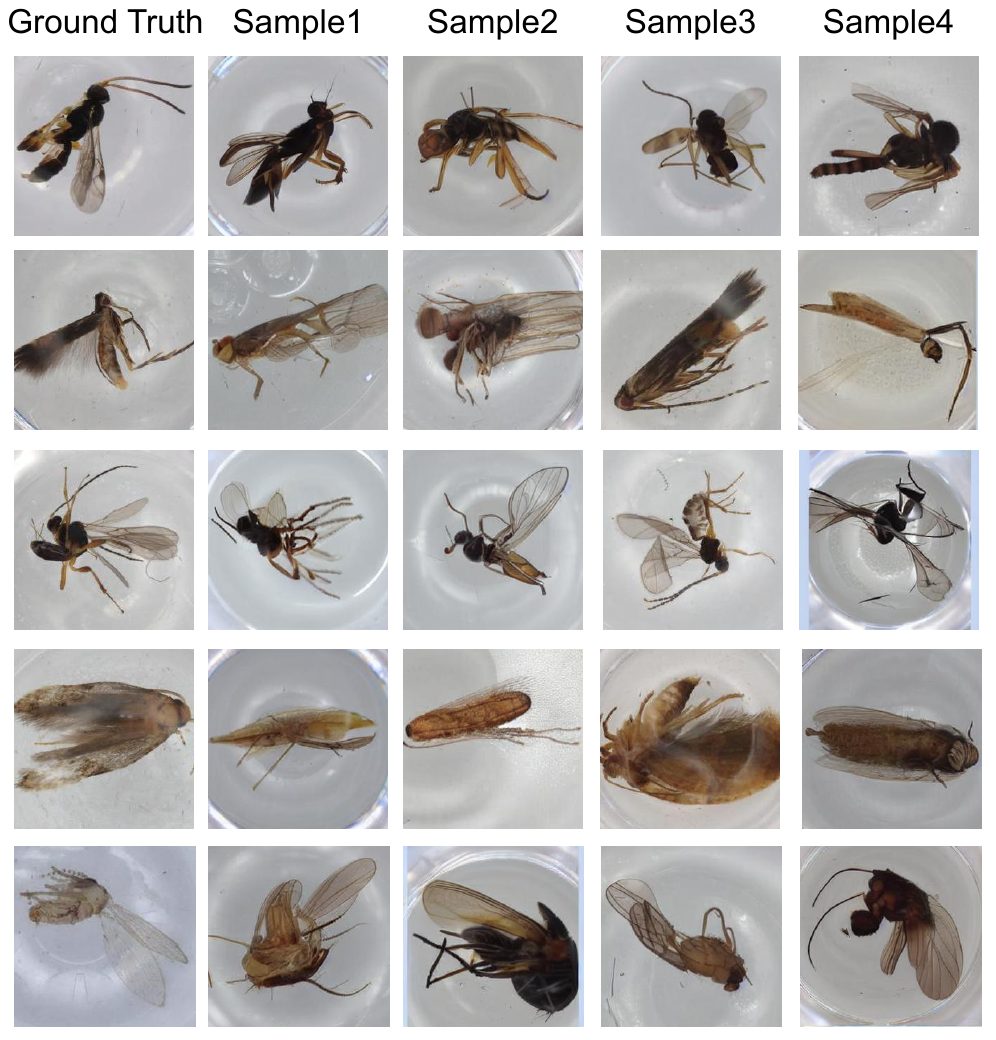}
    \caption{Generative results on unseen species.}
    \vspace{-1.5em}
    \label{fig:unseen}
\end{figure}

Results in Table~\ref{table:Q2_unseen_compare} show that our model maintains high performance on these unseen species, though not as high as on the seen species. 
We show some prediction results for unseen species in Fig.~\ref{fig:unseen}, where most of these predictions can closely match the ground truth phenotypes (the first  three rows). It is an interesting that generative models can produce different view's images for the same species given the same genotype and environment conditions. There are also some predictions that retain the essential traits, although not perfectly match the ground truth. As shown in the last two rows, the model retain key features such as the insect's body color, shape patterns and the overall wing structure.
These findings show the potential of our approach to explore genotype-phenotype relationships, uncover species-specific traits, even in challenging or under-explored species.

\section{Conclusion}
\label{sec:Conclusion}
In this work, we introduce G2PDiffusion, the first diffusion model designed for genotype-to-phenotype image synthesis across multiple species.
We introduce an environment-enhanced DNA encoder and a dynamic aligner.
Experimental results show that our model can predict phenotype from genotype better than baselines.
Notably, we believe this is the pioneering effort to establish a direct pipeline for predicting phenotypes from genotypes through generative modeling, which may open new avenues for research and practical applications in various biological fields.

{
    \small
    \bibliographystyle{ieeenat_fullname}
    \bibliography{main}
}


\end{document}